\documentclass[wcp]{jmlr}

\usepackage{times}
\usepackage{latexsym}
\usepackage{amsfonts}
\usepackage{multirow}
\usepackage{bbding}
\usepackage{color,xcolor}
\usepackage{ulem}
\usepackage{microtype}
\usepackage{inconsolata}
\usepackage{booktabs}
\usepackage{wrapfig}

\usepackage{longtable}

\usepackage{booktabs}

\usepackage{lineno}

\makeatletter
\let\Ginclude@graphics\@org@Ginclude@graphics 
\makeatother

\jmlrvolume{222}
\jmlryear{2023}
\jmlrworkshop{ACML 2023}

\title[Low-Resource Dialogue State Tracking]{State Value Generation with Prompt Learning and Self-Training \\ for Low-Resource Dialogue State Tracking}

  \author{\Name{Ming Gu} \Email{51215901012@stu.ecnu.edu.cn}\\
  \addr School of Computer Science and Technology, East China Normal University
  \AND
  \Name{Yan Yang}\thanks{Corresponding Author} \Email{yanyang@cs.ecnu.edu.cn}\\
  \addr School of Computer Science and Technology, East China Normal University
  \AND
  \Name{Chengcai Chen} \Email{arlenecc@xiaoi.com}\\
  \addr Xiaoi Research, Xiaoi Robot Technology Co., Ltd
  \AND
  \Name{Zhou Yu} \Email{zy2461@columbia.edu}\\
  \addr Dialogue NLP Lab, Columbia University
 }

\editors{Berrin Yan{\i}ko\u{g}lu and Wray Buntine}

\begin{document}

\maketitle

\begin{abstract}
	Recently, low-resource dialogue state tracking (DST) has received increasing attention. First obtaining state values then based on values to generate slot types has made great progress in this task. However, obtaining state values is still an under-studied problem. Existing extraction-based approaches cannot capture values that require the understanding of context and are not generalizable either. To address these issues, we propose a novel \textbf{S}tate \textbf{VA}lue \textbf{G}eneration based framework (\textbf{SVAG}), decomposing DST into state value generation and domain slot generation. Specifically, we propose to generate state values and use self-training to further improve state value generation. Moreover, we design an estimator aiming at detecting incomplete generation and incorrect generation for pseudo-labeled data selection during self-training. Experimental results on the MultiWOZ 2.1 dataset show that our method which has only less than 1 billion parameters achieves state-of-the-art performance under the data ratio settings of 5\%, 10\%, and 25\% when limited to models under 100 billion parameters. Compared to models with more than 100 billion parameters, SVAG still reaches competitive results.\footnote{Our code is available at \url{https://github.com/SLEEPWALKERG/SVAG}}
\end{abstract}

\section{Introduction}
\label{sec-intro}
\begin{wrapfigure}[22]{r}{0.5\textwidth}
	\centering
	\includegraphics[width=0.48\textwidth]{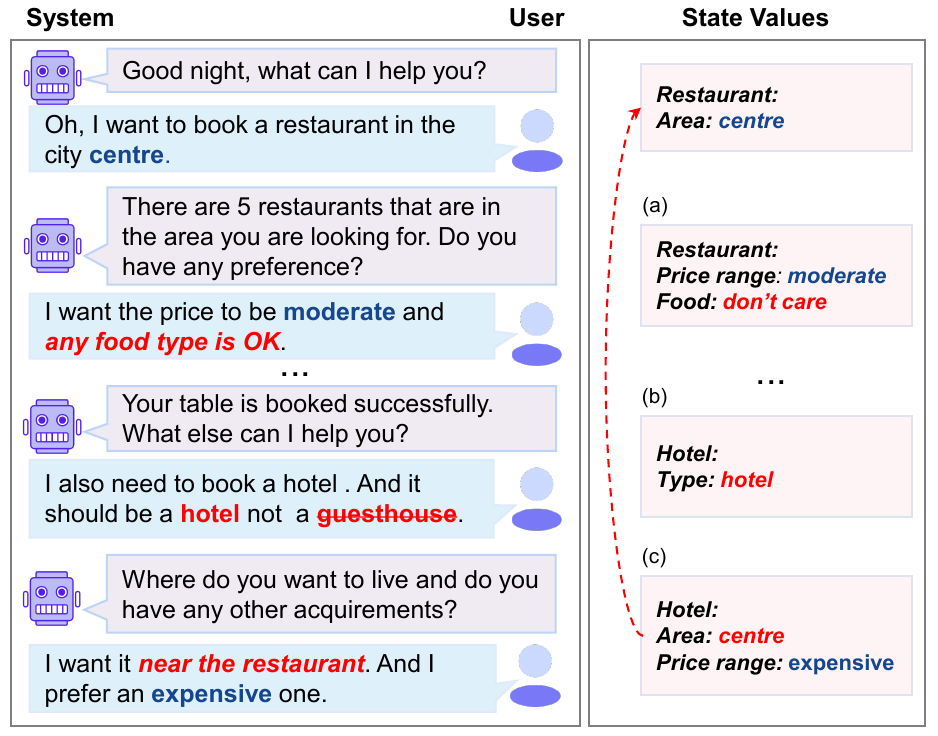}
	\caption{Three main issues of state value generation in DST: (a) \textit{``don't care''} should be generated, (b) \textit{``hotel''} should be distinguished from \textit{``guesthouse''}, and (c) \textit{``centre''} should be inferred from the first turn. Words in blue are state values that can be extracted directly.}
	\label{three-problem}
\end{wrapfigure}

Dialogue State Tracking (DST) is a critical component in task-oriented dialogue systems. It aims to track the dialogue state at every dialogue turn, where the state is represented in forms of a set of (domain-slot, value) pairs. As the rise of new dialogue domains in practice, it poses a big challenge for scenarios with limited resources. Most previous work has attempted to tackle this challenge through cross-domain transfer learning \citep{trade, meta-learning} and cross-task transfer learning \citep{mrc2dst, transfer, ds2}. Moreover, pre-trained language models (PLMs) adaption methods \citep{tod-bert, pptod} are proven to be effective for low-resource DST. However, all these methods suffer from domain or task dependencies. Recently, prompt learning has been applied to low-resource DST by \citet{pl-few-shot-dst}, the method of which demonstrates the potential of prompt learning for generating the slot type of a given state value with limited training data. Therefore, \textit{``first obtain state values, and then generate slot types''} is pointed to be a promising direction for low-resource DST.
The accuracy of state values is critical for the performance of such a two-step method for low-resource DST. However, how to obtain state values correctly is still under explored. Previous approaches \citep{pl-few-shot-dst} simply extract state values with the rule-based method, which significantly limits the accuracy and the generalization. We investigate the state values in DST and find that state value generation has three major issues as shown in Figure \ref{three-problem}. First, some state values may not be extracted directly such as the state value \textit{``don't care''}. A model should understand semantics, then generate these state values. Second, there may be multiple state values appearing in the utterance but only some of them can represent the user's intention. For example, in Figure \ref{three-problem}, \textit{``hotel''} is the state value but the \textit{``guesthouse''} is not. So, a model should have the ability to distinguish them. Third, there are some state values that need to be inferred in context. For example, the state value \textit{``centre''} in hotel booking should be inferred from the dialogue history. Simply extracting this information cannot obtain these correct state values, so we propose not to extract these values but use the generation method to generate correct state values.

Furthermore, since unlabeled dialogue data can be relatively easily obtained in real-world applications, we try to make use of these data to further improve the performance of state value generation. Therefore, we propose to self-train the state value generator to iteratively improve its performance, alleviating the above three issues. For self-training to be effective in the context of generation tasks, it is critical to select confident pseudo-labeled data. Previous methods \citep{qg-self-training, data2text-self-training} use perplexity or some learned metrics to measure the quality of generated sequences. However, state values are not sequences. Therefore, we try to design a state value estimator to estimate the quality of the generated set of state values.

In this paper, we propose \textbf{SVAG}, a \textbf{S}tate \textbf{VA}lue \textbf{G}eneration based framework for low-resource DST with prompt learning and self-training, which decomposes DST into two sub-tasks: state value generation and domain slot generation. Specifically, we first propose a prompt based state value generator, which takes advantage of the PLM to address the three issues above. Second, we propose to self-train the state value generator to further improve its performance and propose a prompt based estimator to filter out noisy pseudo-labeled data during self-training. Finally, a prompt based domain slot generator is proposed to generate the corresponding slot type of a given state value. Experimental results show that SVAG reaches state-of-the-art results on MultiWOZ2.1 under the data ratio settings of 5\%, 10\%, and 25\% when limited to models under 100 billion parameters, demonstrating the superiority of our proposed state value generation based method with self-training. In addition, SVAG also achieves competitive results compared to methods based on models with more than 100 billion parameters.

The contributions of this paper are summarized as the following:
\begin{itemize}
	\itemsep0em
	\item We propose SVAG, an effective and general state value generation based framework for low-resource DST.
	\item We design an estimator with the goal of measuring both the accuracy and completeness of state value generation to filter out noisy pseudo-labeled data during self-training.
	\item Experimental results show that SVAG achieves competitive performance in low-resource DST. 
\end{itemize}

\begin{figure*}[t]
	\begin{center}
		\includegraphics[width=0.8\textwidth]{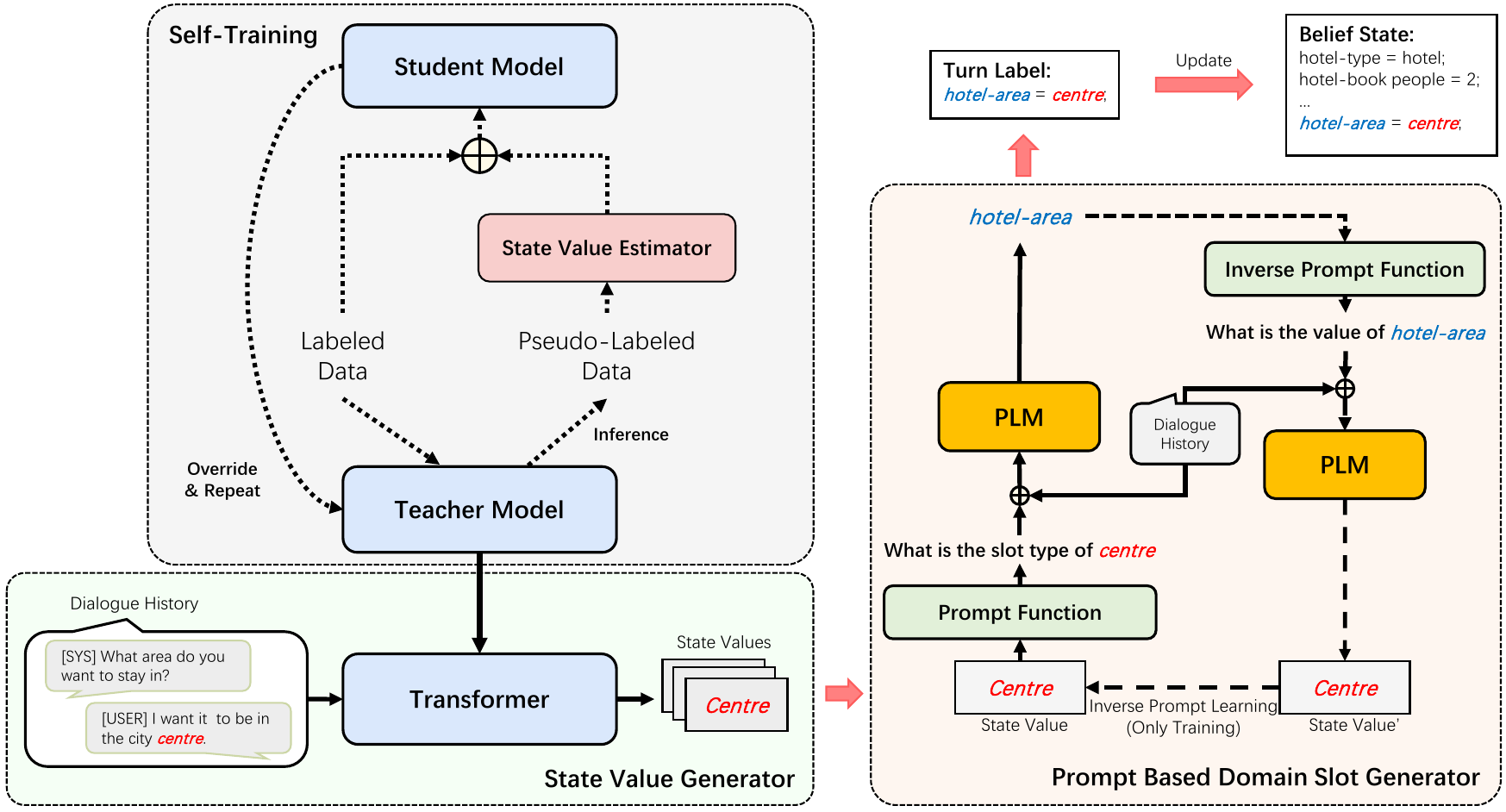}
	\end{center}
	\caption{The overview of our proposed framework. There are three main components of our framework: a state value generator, a self-training strategy, and a domain slot generator. Given the dialogue history, the state value generator first generates the state values in the current turn, then the domain slot generator generates the slot type for each generated state value. Finally, we use the turn labels to update the belief state.}
	\label{framework}
\end{figure*}

\section{Framework}
\label{sec-framework}
In this section, we first set up the notations used throughout the paper. Then we will describe our proposed prompt based framework for low-resource DST which consists of three main components: (1) a state value generator which aims to generate the state values in the current turn; (2) a self-training strategy with a prompt based estimator which aims to boost the performance of the low-resource state value generator with fine-grained selected pseudo-labeled data; (3) a prompt based domain slot generator which aims to generate the corresponding slot type of a given state value. Figure \ref{framework} illustrates the whole framework.

\textbf{Notation.} Let us define $D = \lbrace (S_1, U_1),(S_2, U_2),...,(S_t, U_t) \rbrace$ as the set of system response and user utterance pairs in $T$ turns of a dialogue, where $S_t$ and $U_t$ represent the system's and user's utterance respectively. Also, we define $T = \lbrace T_1, T_2,...,T_t \rbrace$ as the turn label for each turn, where the turn label comprises multiple tuples of domain slots $s$ and their associated values $v$ ($T_t = \lbrace(s_1, v_1), (s_2, v_2),...,(s_n, v_n) \rbrace$). In addition, we define $V_t = \lbrace v_1, v_2,...,v_n\rbrace$ as all the state values in the $t$-th turn label.

\subsection{State Value Generator}
\label{svg}
Given the $t$-th turn utterances $D_t = (S_t, U_t)$ and its history $D_{<t} = (S_{<t}, U_{<t})$, this model aims to generate the state values that the user mentions or confirms at the current turn. Following \cite{dssdst}, we denote the representation of the dialogue history before turn $t$ as $D_{<t} = S_1 \oplus ; \oplus U_1 \oplus ; ... ; \oplus S_{t-1} \oplus ; \oplus U_{t-1}$ and $D_t = S_t \oplus ; \oplus U_t$ is the input of current turn utterances. Finally, the input of the state value generator can be denoted as:
\begin{equation}
	\begin{split}
		X_t = & PREFIX \oplus [HISTORY] \oplus D_{<t} \\
		& \oplus [TURN] \oplus D_t,
	\end{split}
\end{equation}
where $[HISTORY]$ and $[TURN]$ are two special tokens that indicate the start of the dialogue history before turn $t$ and the start of the $t$-th turn utterances, respectively.  $PREFIX$ in our model is \textit{``get the requests that the user confirmed or mentioned in this turn''}. In addition, $\oplus$ is just a simple concatenation operation.

Given this input, a bi-directional transformer \citep{transformer} encoder then outputs:
\begin{equation}
	H_t = Encoder(X_t),
\end{equation}
where $H_t \in \mathbb{R} ^{L \times d}$, $L$ is the length of the input sequence and $d$ is the hidden dimension of the encoder. Then the decoder attends to the encoder output $H_t$ and decodes the corresponding state values. The output is naturally a set of state values in our experiment, but this data structure is not supported by a traditional decoder. So we transfer the output to $V_{output} = v_1 \oplus | \oplus v_2 \oplus | ...  | \oplus v_n$. Then:
\begin{equation}
	\widehat{V}_{output} = Decoder(H_t).
\end{equation}

The overall learning objective of the proposed state value generation processing is to maximize the log-likelihood of $V_{output}$ given the dialogue history before the $t$-th turn $D_{<t}$, the $t$-th turn dialogue $D_{t}$, and the $PREFIX$. That is:
\begin{equation}
	\sum{log P(V_{output} | D_{\leq{t}}, PREFIX)}.
\end{equation}

\subsection{Self-Training with a State Value Estimator}
The process of our self-training approach is shown in Figure \ref{framework}. First, the \textit{Teacher} is applied to generate pseudo labels on unlabeled data $U$. Second, a \textit{Student} is trained on the limited labeled data $S$ and the confident pseudo-labeled data. Lastly, the trained \textit{Student} becomes a new \textit{Teacher}. Multiple iterations are computed till the accuracy no longer increases. The state value generator trained on $S$ acts as the initial \textit{Teacher}. For self-training to be effective in state value generation, it is crucial to carefully select confident pseudo-labeled data to mitigate the risk of reinforcing the model's mistakes. 

We analyze the results of our state value generator and find that it sometimes generates only part of the state values or occasionally generates some incorrect values. Among all the bad cases, the incomplete generation problem occurs most frequently. So, to prevent self-training from further reinforcing the model's mistakes of incomplete generation and incorrect generation, we design a low-resource state value estimator with prompt learning to identify these mistakes as well as filter out these noisy pseudo-labeled data. In order to detect whether the model misses some correct state values or generates some incorrect values, we need a large number of negative samples. We propose synthesizing the examples using the limited labeled dataset.

\begin{figure}[t]
	\centering
	\includegraphics[width=0.48\textwidth]{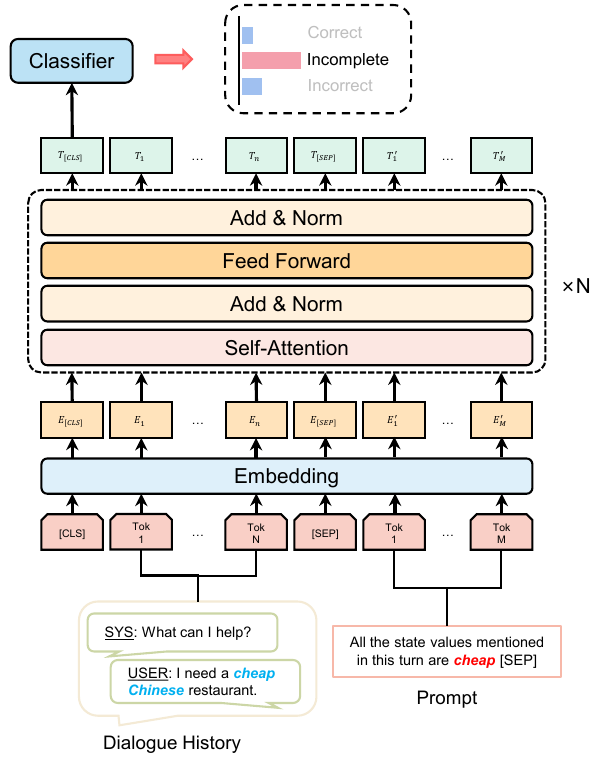}
	\caption{The model architecture of our proposed state value estimator. Given the dialogue history and the generated state values, the model predicts whether all the state values are correctly generated. This figure shows an example that the estimator detects incomplete generation.}
	\label{filter}
\end{figure}
We first describe our proposed state value estimator. Figure \ref{filter} illustrates the model architecture. Given a set of state values $V$, we first convert it to prompt. We manually design a template for the prompt. If there are no state values in the current turn, the prompt is \textit{``there are no values mentioned in this turn.''}. Otherwise, the prompt is \textit{``all the values mentioned in this turn are $v_1,...,v_n$.''}. Then the input of the state value estimator can be denoted as:
\begin{equation}
	\begin{split}
		X_t = & [CLS] \oplus D_{<t} \oplus . \oplus D_t \oplus [SEP] \\
		& \oplus PROMPT \oplus [SEP],
	\end{split}    
\end{equation}
where $[CLS]$ and $[SEP]$ are two special tokens.

Given this input, the output representation of the encoder is $H_t \in \mathbb{R} ^ {|X_t| \times d}$, and $h_t^{[CLS]}$ is the output of that corresponds to $[CLS]$. We then feed $h_t^{[CLS]}$ into an output layer for classification, which can be denoted as:
\begin{equation}
	P = Softmax(MLP(h_t^{[CLS]})),
\end{equation}
where MLP consists of two linear layers and a tanh activation function. In addition, $P \in \mathbb{R}^{|\mathbb{O}|}$ is the probability distribution for each label. In our formulation, $|\mathbb{O}| = 3$, because we set three labels to the estimator. They are (1) correct generation; (2) incomplete generation; (3) incorrect generation. The model is trained with the standard cross-entropy loss.

\begin{figure*}
	\centering
	\includegraphics[width=0.9\textwidth]{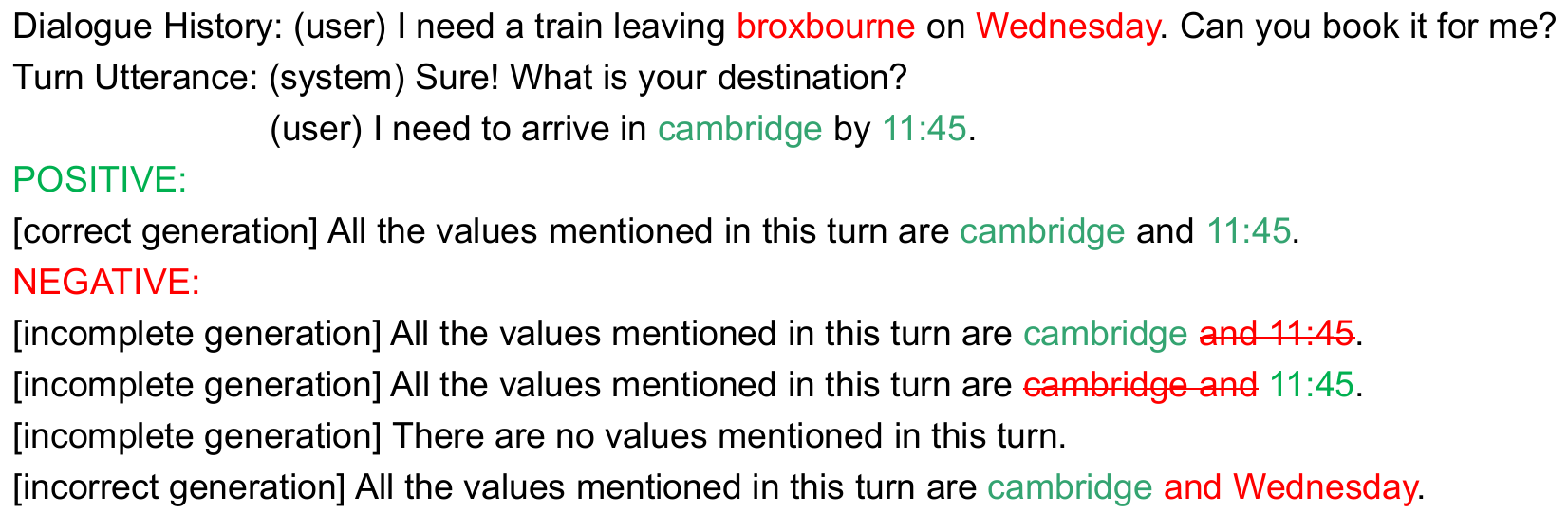}
	\caption{An example of our proposed negative sampling.}
	\label{negative}
\end{figure*}

Finally, we describe how we construct the negative samples. To enhance the model's capability of detecting incomplete generation, which is the key problem of our state value generator, we create much more incomplete generation samples than the others. We randomly remove some values from the ground truth set of state values as incomplete generation samples. Additionally, we add state values that appeared in previous turns to the current state values set as incorrect generation samples. Figure \ref{negative} visualizes an example of our proposed negative sampling. 

\subsection{Prompt Based Domain Slot Generator}
In this section, we describe the domain slot generation method. We use a prompt based model to generate the domain slot. Following \cite{pl-few-shot-dst}, we also add an inverse prompt mechanism to this model which aims to enhance the PLM to better understand the DST task. In our model, we use prompt function $f(v)$ = \textit{``what is the slot type of [v]''} while $I(s)$ = \textit{``what is the value of [s]''} is the inverse prompt \citep{pl-few-shot-dst}. 
Given the state value $v$, we first construct the input of the PLM, which can be denoted as:
\begin{equation}
	\begin{split}
		X_t = & D_{<t} \oplus . \oplus D_t \oplus [SEP] \\
		& \oplus f(v) \oplus [SEP],
	\end{split}
\end{equation}
The overall learning objective of the domain slot generation processing is to maximize the log-likelihood of domain slot $s$ given the dialogue history before the $t$-th turn $D_{<t}$, the $t$-th turn dialogue $D_{t}$, and the value-based prompt $f(v)$. The loss function can be denoted as:
\begin{equation}
	\mathcal{L} = -\sum{log P(s | D_{\leq{t}}, f(v))}.
\end{equation}

Because each turn label $T_t$ may contain multiple $(s, v)$ pairs, each pair of them constructs an instance for training and testing in the domain slot generation model. While testing, the value is generated by our proposed state value generator.

To achieve better domain slot generation performance under low-resource scenarios, we add inverse prompt learning while training like \citet{pl-few-shot-dst}. Just simply replace $f(v)$ with $I(s)$ in $X_t$, then we get the inverse input. Similarly, the loss function for the inverse prompt mechanism is:
\begin{equation}
	\widetilde{\mathcal{L}} = -\sum{log P(v | D_{\leq{t}}, I(s))}.
\end{equation}

Finally, the loss function $\mathcal{L}^*$ consists of both loss functions in prompt learning and inverse prompt learning, which can be denoted as:
\begin{equation}
	\label{w}
	\mathcal{L}^* = \mathcal{L} + w \times \widetilde{\mathcal{L}},
\end{equation}
where $w$ is a weight used to adjust the influence of the inverse prompt learning.

\subsection{Belief State Updating}
Previous sections have described how we get the turn labels. In this section, we describe how we use these turn labels to update the belief state. If the domain-slot-value is not consistent with the previous turns, then we update the corresponding value in the existing belief state. Otherwise, if they don’t have the domain-slot-value tuple, we append it to the existing belief state. For example, in Figure \ref{framework}, \textit{``hotel-area''} doesn't exist in the previous belief state, we just add \textit{``hotel-area-centre''} to it.

\section{Experimentation}
\subsection{Datasets and Metrics}
\textbf{Datasets} We conduct our experiments on the MultiWOZ 2.1 dataset \citep{mwz2-1}. It is a multi-domain task-oriented dialogue dataset which contains 8438 dialogues for training, 1000 dialogues for validating, and 1000 dialogues for testing. Following existing work \citep{trade}, only five domains (restaurant, hotel, attraction, taxi, train) are used in our experiments because the other two domains have very few dialogues and only appear in the training set. 

\textbf{Metrics} The standard metric \citep{trade}, joint goal accuracy (JGA) is used in our experiments. This metric compares the whole predicted belief state to the gold one at each dialogue turn. If and only if all the predicted states match the ground truth states exactly for all domains, the prediction is treated as correct. In addition, we use a turn level accuracy (TLA) metric to evaluate the performance of state value generation. This metric compares the predicted state values to the gold ones. Only if all the predicted state values match the ground truth values, the prediction is considered correct.

\subsection{Implementation Details}
We implement the state value generator based on T5 \citep{t5} as well as the domain slot generator and the state value estimator is based on RoBERTa \citep{roberta}. We use the pre-trained checkpoint from \textit{transformers} library\footnote{\url{https://huggingface.co/t5-large}; \url{https://huggingface.co/roberta-base}}. Additionally, we also use \textit{pytorch lightning} library\footnote{\url{https://www.pytorchlightning.ai}} to implement our framework.  All models are trained using the AdamW \citep{adamw} optimizer with a linear learning rate decay. The peak learning rate of the two generators is 5e-5 while 2e-5 is for the state value estimator.  We conduct our experiments under the data ratio setting of 1\%, 5\%, 10\%, and 25\%.  For domain slot generation, we set $w$ in Eq \ref{w} to 0.1 \citep{pl-few-shot-dst}.

To select the best checkpoint of the state value estimator for pseudo-labeled data selection, we validate the model with our synthesized validation dataset. The F1-score of correct generation examples is used to choose the best model. We choose the threshold of our state value estimator based on the blank rate\footnote{blank here means that there are no state values that should be generated at the current turn} of the pseudo-labeled data as it gets higher when the threshold goes up. Finally, the threshold of the estimator is set to $0.98$.

\begin{table*}[t]
	\centering
	\small
	\resizebox{\linewidth}{!}{
		\begin{tabular}{c|c|c|c|c|c|c|c}
			\toprule
			\multirow{2}{*}{\textbf{Model}} & \textbf{Param.} & \textbf{Unlabeled}& \textbf{External} &
			\multicolumn{4}{|c}{\textbf{Data ratio}} \\
			\cmidrule{5-8}
			& \textbf{Size} & \textbf{MultiWOZ Data} & \textbf{Data} & 1\% & 5\% & 10\% & 25\% \\
			\midrule
			\textbf{TRADE} \citep{self-sup} & \multirow{14}{*}{\textless1B} & No & - & 10.4 & 27.7 & 32.6 & 38.5 \\
			\textbf{BERT} \citep{tod-bert} & & No & - & 6.4 & 19.6 & 32.9 & 40.8 \\
			\textbf{BERT}$^{*}$ \citep{st-tod} & & No & - & 8.0 & - & 21.2 & - \\
			\textbf{SGPDST} \citep{sgpdst} & & No & - & 32.1 & 43.1 & 46.9 & - \\
			\textbf{TOD-BERT} \citep{tod-bert} & & No & ToD Data & 9.9 & 28.6 & 39.5 & 44.3 \\
			\textbf{TOD-BERT}$^{*}$ \citep{st-tod} & & No & ToD Data & 8.4 & - & 25.5 & - \\
			\textbf{DS2} \citep{ds2} & & No & Summary Data & 33.8 & 44.2 & 45.4 & - \\
			\textbf{PL-DST} \citep{pl-few-shot-dst} & & No & Labeled ToD Data & \textbf{44.3} & 44.7 & 44.7 & 45.4 \\
			\textbf{Self-Sup} \citep{self-sup} & & Yes & - & 19.5 & 30.6 & 34.5 & 40.2 \\
			\textbf{BERT-ST} \citep{st-tod} & & Yes & - & 8.8 & - & 23.9 & - \\
			\textbf{TOD-BERT-ST} \citep{st-tod} & & Yes & ToD Data & 9.9 & - & 28.3 & - \\
			\textbf{SVAG (Our Method)} & & Yes & - & 31.9 & \textbf{45.1} & \textbf{47.6} & \textbf{50.0} \\
			\textbf{SVAG w/o ST} & & No & - & 31.9 & 43.5 & 44.6 & 48.2 \\
			\midrule
			\textbf{IC-DST GPT-Neo 2.7B} \citep{ic-dst} & \multirow{4}{*}{\textless100B} & No & - & 16.7 & 26.9 & 31.7 & - \\
			\textbf{IC-DST CodeGen 2.7B} \citep{ic-dst} & & No & - & 20.7 & 29.6 & 33.8 & - \\
			\textbf{SM2-3B} \citep{sm2} & & No & Labeled ToD Data & 38.1 & 39.9 & 39.9 & - \\
			\textbf{SM2-11B} \citep{sm2} & & No & Labeled ToD Data & 38.4 & 44.6 & 46.0 & - \\
			\midrule
			\textbf{IC-DST CodeX-davinc 175B} \citep{ic-dst} & \textgreater100B & No & - & 43.1 & 47.1 & 48.7 & - \\
			\bottomrule
		\end{tabular}
	}
	\caption{\label{main-result}
		Comparison of different models' JGA scores for low-resource DST on MultiWOZ 2.1 \citep{mwz2-1} under different ratios of training data. We report the averaged JGA score of SVAG over three runs. Bolded numbers indicate highest performance on models under 1 billion parameters. *: taken from \cite{st-tod}.
	}
\end{table*}

\subsection{Baseline Models}
We compare our proposed method with several strong baselines for low-resource DST.

\textbf{TRADE} \citep{self-sup} uses a Seq2Seq model to decode the corresponding value for each predefined slot with a soft copy mechanism. It is trained with limited MultiWOZ data.

\textbf{BERT} \citep{tod-bert} treats DST as a multi-class classification problem using a predefined ontology and is trained with limited MultiWOZ data.

\textbf{SGPDST} \citep{sgpdst} uses domain, slot, and slot description as prompt and fine-tunes a PLM to generate the corresponding value with limited MultiWOZ data.

\textbf{TOD-BERT} \citep{tod-bert} continues BERT's pre-training on several external task-oriented dialogue (ToD) datasets, then adopts the model to low-resource DST like BERT \citep{tod-bert}.

\textbf{DS2} \citep{ds2} uses a template based method to convert dialogue states to summaries and reformulate DST as dialogue summary. They first fine-tune the model with dialogue summary datasets and then fine-tune it with limited MultiWOZ data.

\textbf{PL-DST} \citep{pl-few-shot-dst} uses a prompt based method to generate the slot type for a given state value with limited MultiWOZ data. Their model is based on SOLOIST which is pre-trained with several external ToD datasets and adds DST to pre-training tasks. Moreover, their model ignores domain information and might use the ground truth domain while evaluating.

\textbf{Self-Sup} \citep{self-sup} adds two self-supervised objectives for TRADE using the rest of unlabeled MultiWOZ data to improve its performance in low-resource scenarios.

\textbf{BERT-ST} \citep{st-tod} proposes a text augmentation technique for the limited MultiWOZ data and then self-trains BERT \citep{tod-bert} by using the rest of unlabeled MultiWOZ data and the augmented data.

\textbf{TOD-BERT-ST} \citep{st-tod} proposes the same method as BERT-ST \citep{st-tod} on top of TOD-BERT. So it not only utilizes the rest of unlabeled MultiWOZ data but also uses external ToD data for pre-training.

\textbf{IC-DST} \citep{ic-dst} reformulates DST as a text-to-SQL task and uses in-context learning to prompt a CodeX model with limited MultiWOZ data.

\textbf{SM2} \citep{sm2} stabilizes in-context learning by using meta-learning with external labeled ToD data and then leverages in-context learning to DST with limited MultiWOZ data.

\begin{table}[t]
	\small
	\centering
	\begin{tabular}{c|c|c|c|c}
		\toprule
		\textbf{Data} & \textbf{ST-} & \textbf{No. of} & \multirow{2}{*}{\textbf{TLA}} & \multirow{2}{*}{\textbf{JGA}} \\
		\textbf{ratio}	& \textbf{iter.} & \textbf{samples} & & \\
		\midrule[0.5pt]
		\multirow{3}{*}{1\%} & - & 578 & 66.71\% & \textbf{30.77\%} \\
		& 1 & +30,852 & \textbf{67.28\%} & 30.55\% \\
		& 2 & +5,963 & 66.22\% & 29.03\% \\
		\midrule
		\multirow{3}{*}{5\%} & - & 2,807 & 74.20\% & 42.43\% \\
		& 1 & +38,918 & \textbf{76.13\%} & \textbf{45.06\%} \\
		& 2 & +2339 & 75.65\% & 44.29\% \\
		\midrule
		\multirow{3}{*}{10\%} & - & 5,626 & 75.37\% & 43.24\% \\
		& 1 & +28,956 & \textbf{77.66\%} & \textbf{47.48\%} \\
		& 2 & +2,482 & 77.65\% & 47.27\% \\
		\midrule
		\multirow{3}{*}{25\%} & - & 13,932 & 78.23\% & 49.02\% \\
		& 1 & +20,334 & \textbf{78.95\%} & \textbf{50.61\%} \\
		& 2 & +1,255 & 78.60\% & 49.62\% \\
		\midrule
		Full data & - & 56668 & 79.87\% & 51.41\% \\
		\bottomrule
	\end{tabular}
	\caption{\label{self-training}
		Model performance over multiple self-training iterations under different data ratio settings. \textit{``ST-iter.''} denotes the iteration of self-training.
	}
\end{table}

\subsection{Main Results}
Following the previous work \citep{trade}, we randomly select limited labeled data from the training set to simulate the low-resource scenarios with three different random seeds (10, 20, and 48). We conduct our experiments using 1\%, 5\%, 10\%, and 25\% data. Note that 1\% data has only 84 dialogues. Table \ref{main-result} shows the JGA score of our method and other baselines in different data ratio settings. Among the models using unlabeled MultiWOZ data, our framework SVAG achieves state-of-the-art performance.
\begin{table}
	\centering
	\small
	\begin{tabular}{p{0.95\columnwidth}}
		\toprule
		\textbf{Dialogue history:} [user] I would like a reservation for 2 to the Peking restaurant.  \\
		\textbf{Current Turn Utterances:} [sys] OK, and what day and time would you like that reservation?
		
		[user] I would like to make a reservation for {\color[RGB]{84, 180, 53}{Saturday}} at {\color[RGB]{84, 180, 53}{11:45}}. And there has been a change in plans, I will be dining {\color[RGB]{84, 180, 53}{alone}}. \\
		\textbf{Without-ST prediction:} Saturday, 11:45 \\
		\textbf{With-ST prediction:} Saturday, 11:45, \textbf{1} \\
		\textbf{Ground truth: } Saturday, 11:45, 1 \\
		\midrule
		\textbf{Dialogue history:} [user] I am staying in Cambridge soon and would like to stay at  {\color[RGB]{84, 180, 53}{a and b guest house}}. ... [sys] Your booking is successful! ... [sys] there are actually 7 museums in that area. [user] Great, can I get the postcode, entrance fee and address of 1 of them? \\
		\textbf{Current Turn Utterances:} [sys]  {\color[RGB]{84, 180, 53}{cafe jello gallery}} has a free entrance fee. The address is cafe jello gallery, 13 magdalene street and the post code is cb30af. Can I help you with anything else? 
		
		[user] Yes please. I need a taxi to {\color[RGB]{84, 180, 53}{commute}}. \\
		\textbf{Without-ST prediction:} cafe jello gallery \\
		\textbf{With-ST prediction:} cafe jello gallery, \textbf{a and b guest house}  \\
		\textbf{Ground truth:} a and b guest house, cafe jello gallery \\
		\bottomrule
	\end{tabular}
	\caption{\label{examples-st}
		Examples that self-training outperforms without self-training in state value generation.}
\end{table}
We observe that SVAG has a measurable improvement over TRADE \citep{self-sup} and BERT \citep{tod-bert}. SVAG also outperforms TOD-BERT \citep{tod-bert}, DS2 \citep{ds2}, and PL-DST \citep{pl-few-shot-dst} under the data ratio setting of 5\%, 10\%, and 25\%, although all of them are enhanced by external data.

In particular, we observe that SGPDST \citep{sgpdst} and DS2 \citep{ds2} achieves a higher JGA score than ours under the data ratio setting of 1\%. SGPDST \citep{sgpdst} achieves better performance because it not only uses domain slot information but also leverages slot description to facilitate the low-resource DST, making the model better understand the task. However, it is costly to collect all these information. Our model doesn't rely on a given schema. DST is naturally a process of summarizing important information. Therefore, DS2 \citep{ds2} achieves great performance in extreme low-resource scenarios because it reformulates DST as dialogue summary with rule-based templates from dialogue states and pre-trains the model on external summary data. However, their model cannot be applied to general scenarios since such great performance comes from the manual rules and the extra dialogue summary data. Additionally, constructing manual rules for new domains is costly. Conversely, our method does not rely on any other annotation, providing a more general and efficient solution for low-resource DST.

We also observe that PL-DST \citep{pl-few-shot-dst} achieves the best result under the data ratio setting of 1\% because it is based on SOLOIST \citep{soloist} which is pre-trained on several external ToD datasets and adds DST to pre-training tasks. Furthermore, their framework also ignores domain information and might use the ground truth domain while evaluating, so it is unfair to compare it to all the other methods. Domain prediction is actually not that easy if there are close-by domains that share the same slot types, such as in MultiWOZ. In addition, it simply extracts state values with a rule-based method, which significantly limits their model's generalization. We observe that the more data, the better SVAG is, which demonstrates the effectiveness of our method.

Moreover, we observe that SVAG significantly outperforms Self-Sup \citep{self-sup} and BERT-ST \citep{st-tod} with a great margin under all data ratio settings. These two models are comparable to our method since both of them use unlabeled MultiWOZ data and don’t use any external data. The higher JGA scores of SVAG demonstrate not only the superiority of our proposed state value generation based method for low-resource DST but also the efficiency of our proposed state value estimator based self-training for utilizing unlabeled MultiWOZ data. We also observe that SVAG outperforms TOD-BERT-ST \citep{st-tod} which not only uses unlabeled MultiWOZ data but also incorporates external ToD data into pre-training, indicating that SVAG can make better use of the PLM's strong capability of comprehension to fulfill the DST task.

Recently, as the rise of large language models (LLM), many in-context learning based approaches have been proposed for DST. Although SVAG has less than 1B parameters, SVAG still outperforms SM2 (3B \& 11B)\citep{sm2} under the data ratio settings of 5\% and 10\%. Additionally, SVAG outperforms IC-DST (\textless100B) \citep{ic-dst} under all data ratio settings. SM2 achieves a much higher JGA socre than ours under the data ratio setting of 1\% because it uses several labeled ToD datasets for meta-learning and the base model of it is much bigger than ours. Moreover, SVAG also achieve comparable performance compared to IC-DST(175B) \citep{ic-dst} under the data ratio setting of 5\% and 10\%, although its parameters are more than 100 times ours.

\begin{table}
	\centering
	\small
	\begin{tabular}{p{0.95\columnwidth}}
		\toprule
		\textbf{Dialogue history:} [sys] {\color[RGB]{84, 180, 53}{Saigon city}} is Asian oriental and is in the north as the hotel is. It is expensive. Shall I book it for you ? [user] Yes, please . I need reservations for {\color[RGB]{84, 180, 53}{12:30}} on {\color[RGB]{84, 180, 53}{Friday}}. There are {\color[RGB]{84, 180, 53}{5}} in my group. \\
		\textbf{Generated state values:} Friday, 5, 12:30 \\
		\textbf{Ground truth:} Friday, 5, 12:30, \textbf{Saigon city} \\
		\textbf{Correct Score:} 0.08  \\
		\textbf{Incomplete Score:} 0.92 \color{green}\checkmark \\
		\textbf{Incorrect Score:} 0.00  \\
		\midrule
		\textbf{Dialogue history:} [user] I would like to visit a park on the north side. [sys] Sure , we have milton country park located in the north in milton. ... [user] OK, great thanks. I also need to find a train going to {\color[RGB]{84, 180, 53}{Cambridge}} \\
		\textbf{Generated state values:} \color{red}milton country park \\
		\textbf{Ground truth:} Cambridge \\
		\textbf{Correct Score:} 0.03  \\
		\textbf{Incomplete Score:} 0.02 \\
		\textbf{Incorrect Score:} 0.95 \color{green}\checkmark \\
		\bottomrule
	\end{tabular}
	\caption{\label{examples-estimator}
		Examples that our proposed state value estimator correctly detects different errors in state value generation. The three scores are the estimator's predicted probability of the three types of generation.}
\end{table}
\subsection{Effectiveness of Self-Training}
\label{effectiveness-st}
In this section, we will analyze the effectiveness of self-training for low-resource state value generation. We run 2 iterations of self-training in our experiment. Table \ref{self-training} reports the performance of our state value estimator based self-training under different data ratio settings. We observe that a significant improvement on TLA is made in the first iteration. We then analyze the pseudo-labeled data selected by the estimator and find that both the quantity and the quality of pseudo-labeled data with high confidence in the first iteration are high. In the second iteration, pseudo-labeled data with high confidence is greatly reduced but noisy examples are increased, resulting in a little decrease in the performance. In addition, we observe the performance of our estimator gets better with the increase of data ratio settings, which makes the performance degradation less obvious.

In the last line of table \ref{self-training}, we also report the results under the full data setting. We observe that our method achieves a close performance over the full data setting when 25\% data is available, which indicates the effectiveness of our proposed state value generator with a state value estimator based self-training strategy. Specially, under the data ratio setting of 1\%, we observe that JGA is not proportional to TLA. It is because we update the belief state with the turn labels. The earlier the error occurs, the greater the impact on JGA due to the error accumulation.

Table~\ref{examples-st} shows two examples of our proposed state value generator with self-training. In the first example, the user briefly informs the system that he/she will book the table for one person by saying \textit{“alone”}. The model without self-training misses the state value \textit{“1”}. After self-training, the model can better understand semantics and generate it. In the second example, the user informs that he/she needs a Taxi to commute from the guest house to the attraction by saying \textit{“I need a Taxi to commute”}. The model without self-training misses the guest house, while the model after self-training can better infer information through context. To sum up, our proposed state value estimator based self-training can significantly enhance the model’s capability of natural language understanding (NLU) and generate the correct state values in low-resource scenarios.

\subsection{Effectiveness of the State Value Estimator}
\begin{wraptable}{r}{0.5\textwidth}
	\centering
	\begin{tabular}{c|c|c}
		\toprule
		\textbf{Data} & \textbf{Selection} & \multirow{2}{*}{\textbf{TLA}} \\
		\textbf{ratio} & \textbf{strategy} \\
		\midrule
		\multirow{3}{*}{1\%} & - & 66.71\%  \\
		& vanilla & 67.25\%  \\
		& value estimator & \textbf{67.28\%} \\
		\midrule
		\multirow{3}{*}{5\%} & - & 74.10\%  \\
		& vanilla & 75.72\%  \\
		& value estimator & \textbf{76.13\%} \\
		\midrule
		\multirow{3}{*}{10\%} & - & 75.37\%  \\
		& vanilla & 76.23\%  \\
		& value estimator & \textbf{77.66\%} \\
		\midrule
		\multirow{3}{*}{25\%} & - & 78.23\%  \\
		& vanilla & 78.80\%  \\
		& value estimator & \textbf{78.95\%} \\
		\bottomrule
	\end{tabular}
	\caption{\label{value checker}
		Comparing performance in terms of Turn Level Accuracy between vanilla and state value estimator based pseudo state values selection strategies. Selection strategy \textit{``-''} denotes training without self-training.
	}
\end{wraptable}
In this section, we will analyze the effectiveness of our proposed state value estimator. Table \ref{examples-estimator} shows two examples of the state value estimator, in which the estimator correctly identifies incomplete generation and incorrect generation. It demonstrates that our proposed state value estimator correctly distinguishes incomplete generation from correct generation, which mitigates the risk of reinforcing incomplete generation for the state value generator.

We also compare the performance of the state value estimator based pseudo state values selection with that of \textit{vanilla} self-training. For each experiment, we \textit{randomly} sample an equal number of examples for vanilla self-training. Table \ref{value checker} summarizes the results. We observe that the state value estimator based self-training improves over vanilla self-training under all data settings. It not only demonstrates the effectiveness of our proposed state value estimator based self-training but also confirms that our negative sampling based synthesized dataset is effective to train the estimator with the purpose of evaluating both the accuracy and completeness.

\section{Related Work}
\subsection{Low-Resource Dialogue State Tracking}
Various approaches have been proposed to low-resource DST. One line of these methods is cross-domain transfer \citep{trade, meta-learning}, which aims at transferring knowledge from one domain to the others. The reason why these methods do work is that many domains share a lot of common slots. The other line of work can be summarized as cross-task transfer \citep{mrc2dst, transfer, ds2}. These methods try to leverage data from another task to facilitate the low-resource DST. For example, \citet{mrc2dst} modeled DST as machine reading comprehension (MRC). They first pre-trained a model on MRC data, then further trained the model with DST data. In addition, PLM adaption methods have been proven to be efficient for low-resource DST. \citet{tod-bert} continued BERT's pre-training on several ToD datasets, then adopted the obtained TOD-BERT to DST task. However, all these methods suffer from domain or task dependencies. Additionally, various in-context learning based methods \citep{ic-dst, sm2} have been proposed to leverage LLMs to low-resource DST. However, these methods are highly dependent on the inference language model and the cost of inference is much higher too. Recently, \citet{pl-few-shot-dst} introduced a prompt based slot generation framework for low-resource DST. However, they obtained state values by a rule-based method, which limits their model's generalization. Different from these methods, we propose a state value generation based method for low-resource DST.

\subsection{Pseudo-Labeled Data Selection in Self-Training}
For self-training, how to deal with noisy, low-quality pseudo labels is crucial to the final performance. Self-training is originally designed for classification problems \citep{revisit-self-training-nlg}. It is easy for these classification models to filter out noisy labels by their `confidence', which is the predicted probability of the label. For self-training in NLG, confidence estimation has been defined as the task of evaluating the quality of the whole sequence of words in the target sentence \citep{qg-self-training}. \citet{qg-self-training} tried to use both sentence perplexity and the BERT-based fluency score to represent the quality. \citet{data2text-self-training} repurposed BLEURT to be a quality estimator. However, our work aims to evaluate the quality of a set of generated state values, rather than the sequences. We propose a prompt based estimator for measuring both the accuracy and completeness of the generated set of state values.

\section{Conclusion and Future Work}
In this paper, we propose SVAG, a prompt based framework for low-resource DST which consists of a state value generator and a domain slot generator. We reveal three issues in state value generation and propose a prompt based state value generator to alleviate them. In order to make use of large amounts of unlabeled dialogue data, we propose to self-train the state value generator. In addition, a state value estimator is designed to filter out noisy pseudo-labeled data during self-training. Moreover, we synthetically generate datasets for training the estimator with the goal of detecting incomplete generation and incorrect generation. Experimental results on MultiWOZ2.1 illustrate the superiority of SVAG over previous approaches for low-resource DST.

In the future work, we plan to mitigate exposure bias in state value generation to alleviate the problem of incomplete generation.

\acks{This work was supported by the Science and Technology Commission of Shanghai Municipality (No. 22511105901, 20511101205) and Shanghai Chinafortune Co.,Ltd. We would like to thank all the anonymous reviewers for their kind comments.}

\bibliography{acml23}
%
%
%
%
%

\end{document}